\documentclass[11pt]{article}
\usepackage[margin=1in]{geometry}
\usepackage{graphicx}
\usepackage{fancyvrb}
\usepackage{hyperref}
\usepackage{svg}
\title{Grammars and reinforcement learning for molecule optimization}
\author{Egor Kraev}

\begin{document}

\maketitle

\begin{abstract}
We seek to automate the design of molecules based on specific chemical properties. Our primary contributions are a simpler method for generating SMILES strings guaranteed to be chemically valid, using a combination of a new context-free grammar for SMILES and additional masking logic; and casting the molecular property optimization as a reinforcement learning problem, specifically best-of-batch policy gradient applied to a Transformer model architecture.

This approach uses substantially fewer model steps per atom than earlier approaches, thus enabling generation of larger molecules, and beats previous state-of-the art baselines by a significant margin.

Applying reinforcement learning to a combination of a custom context-free grammar with additional masking to enforce non-local constraints is applicable to any optimization of a graph structure under a mixture of local and nonlocal constraints.\footnote{The source code to produce these results can be found at \emph{https://github.com/ZmeiGorynych/generative\_playground} }

\end{abstract}
\section{Introduction}
An important challenge in drug discovery is to find molecules with desired chemical properties. While ultimate usefulness as a drug can only be determined in a laboratory or clinical context, that process is expensive, and it is thus advantageous to pre-select likely candidates in software. 

While deep learning has been extensively investigated for molecular graph encoding (\cite{Duvenaud15}, \cite{Kearnes16}, \cite{Gilmer17}), molecule generation is still subject of active research. The simplest natural approach to candidate molecule generation is to generate some sort of a linear representation, such as a string of characters in the SMILES format \cite{Weininger88}, using an encoder-decoder network architecture similar to that used in machine translation, as done in \cite{Gomez-Bombarelli16}.

This approach's performance was comparatively poor because a molecule's structure is not linear, but rather a graph which typically includes cycles, so it falls to the model to learn how to generate SMILES strings that correspond to chemically valid molecules - a nontrivial task that leaves the model with little spare capacity to additionally optimize a given chemical metric of the molecules produced.

A way to partially remedy that involves generating not the actual SMILES strings, but a sequence of production rules of a context-free grammar (CFG) for SMILES, as done by \cite{kusner17}. That guarantees that the SMILES strings produced are grammatically valid, putting less burden on the model to ensure validity and thereby achieving better metrics. 

However, \cite{kusner17} give two reasons why this is still not guaranteed to produce \emph{chemically} valid molecules: firstly, a grammatically valid SMILES string is not guaranteed to be chemically possible (because of atom valences being wrong, for example), and secondly, because a model typically assumes a maximum number of steps per molecule, and the rule expansion is not guranteed to terminate by that number of steps. 

Rather than generating SMILES strings, one could directly generate a molecular graph out of a vocabulary of valid components, an approach taken by \cite{jin18}. An important insight of that approach is that one can regard a molecule as an acyclic graph if one considers cycles to be graph nodes, along with atoms not part of a cycle. Thus, by first constructing such a graph and then resolving that into an actual molecule, the molecules produced by this approach are guaranteed to be chemically valid, so model optimization can focus on optimizing the properties of the actual molecule, resulting in state-of-the-art scores.

Unfortunately the approach of \cite{jin18} is also quite complex, using a two-part internal latent representation, a graph message-passing network, a tree message-passing network, and scoring for subgraphs as they are being assembled.  Information transfer between nodes was done via message passing along the graph edges using an RNN, with the information distance between nodes thus being dependent on the graph traversal order. Furthermore, it seems to require every possible cycle we generate to be contained in a predefined vocabulary, leading to potentially huge vocabularies, especially if we wish to take into account features such as chirality and stereoisomers.
 
Our major contribution is an approach that also guarantees the chemical validity of any molecule produced, but is simpler, more expressive, and more memory-efficient, and results in scores substantially exceeding those of \cite{jin18}.
 
We consider that a context free grammar (CFG) is a very natural way of representing an acyclic graph. We formulate a new grammar for SMILES strings that guarantees correct atom valences by construction, and allows for arbitrarily long, branching chains of aliphatic cycles, and the more common kinds of aromatic cycle structures. Due to the recursive nature of a CFG, we can represent a combinatorially large number of possible cycle structures with a small number of expansion rules (for example, an aliphatic 5-cycle can consist of any 5 sub-structures with 2 free valences each, some of which can also form part of other cycles).
 
To address the issues identified by \cite{kusner17}  we introduce additional masking into the process of selecting the next production rule. One kind of masking tracks the `terminal distance' of the current sequence, making sure the rule expansion terminates before the maximum number of steps allowed by the model. 
 
Another kind of masking makes sure all cycles are closed, and enforces minimum and (optionally) maximum cycle length (cycles of length 2 are illegal in the SMILES format). We achieve that by attaching additional information, such as a unique cycle ID, to any grammar token that indicates creation of a new cycle, and propagating it when expanding that token.
 
All the molecules produced using that additional masking are still valid according to our CFG, so we can use standard CFG parsers (for example, \verb|nltk| \cite{BirdKleinLoper09NLTK}) to decompose the molecules from the ZINC database \cite{Weininger88} into the production rules of our grammar.
 
As discussed in Section~\ref{sec:expressiveness}, our grammar is only expressive enough to represent a little over a third of the molecules in the ZINC database, not due to any inherent limitation of the approach, but merely because that is sufficient to represent the optimized molecules from the literature, eg those in \cite{kusner17} and \cite{jin18}, and to generate new molecules with scores exceeding state of the art. 
 
The second innovation we make is that instead of using an RNN, be it over the sequence of rules, or via graph message-passing, we base our neural model on the Transformer architecture \cite{Transformer17}. The Transformer's information distance between any two inputs is always one, so we have the full information about the sequence so far at our immediate disposal when expanding the next rule. As the memory requirements of the Transformer grow approximately as a square of the number of generation steps, it helps that the new grammar gives a concise representation of a molecule (with approximately half as many rules needed per atom as in the grammar used in \cite{kusner17}, see Section~\ref{sec:expressiveness})
 
 We could have used this architecture in an autoencoder, followed by Bayesian optimization, as has been done in the literature so far. However, as the properties of that approach are well known by now, we decide to try a different tack, namely consider the molecule generation problem as a reinforcement learning problem, and optimize using best-of-batch policy gradient. 
 
 We first train a baseline model off-policy to teach it the distribution of molecules in the ZINC database, and then conduct a series of optimization experiments. Using best-of-batch policy gradient, with varying degrees of anchoring to the baseline model, demonstrates a smooth tradeoff between objective optimization and similarity to the molecules in the ZINC dataset, and yields scores exceeding state of the art even with the more conservative parameter settings.

\section{Generating guaranteed valid SMILES strings}\label{sec:valid_smiles}
\subsection{SMILES}
SMILES \cite{Weininger88} is a common format for describing molecules with a character string.
 
Briefly, atoms following each other in the string are assumed to be connected with a single bond, double bonds are denoted with \verb|=|, and triple bonds with \verb|#|, thus for example $CO_2$ is represented as \verb|O=C=O| and $N_2$ as \verb|N#N|. Side branches are represented by brackets, thus $SO_3$ is \verb|O=S(=O)=O|, aromatic atoms (those contributing one valence to an aromatic ring) are represented as lowercase letters, and cycles are represented by numerals following an atom, with atoms being followed by the same numeral understood to be connected; thus for example benzene is \verb|c1ccccc1|. Valences not used in the explicitly specified bonds are assumed to have hydrogens attached, for example $H_2O$ is \verb|O|.

Existing molecule databases, such as the ZINC database \cite{zinc15} we use, are in SMILES format, and can be processed by software such as \verb|http://www.rdkit.org| \cite{rdkit}.
\subsection{Context-free grammar}
Non-binary trees are a natural description of molecular structure, be it using the SMILES approach (cutting the cycles), or the approach of \cite{jin18} (treating cycles as graph nodes to be resolved into individual atoms at a later stage).

A context-free grammar (CFG) is, in turn, a natural way of describing non-binary trees. A CFG is a set of tokens, and a set of production rules, each rule mapping a token (the rule's `left hand side') onto a sequence of tokens (the rule's `right hand side'). Tokens that occur on the left hand side of at least one production rule are called nonterminals, the other tokens terminals.

A sequence that's valid according to a CFG is generated by starting with the sequence consisting of the root token, and then repeatedly taking a (typically the leftmost) nonterminal token in the sequence, selecting a rule whose left hand side equals that token, and replacing the token with the right hand side of that rule. That process terminates once there are no nonterminals left in the sequence. If we consider that process, not as replacing the token in question, but as inserting the new tokens as its children, the process of generating the sequence is equivalent to generating a non-binary tree whose nodes are nonterminals and whose leaves are terminals. That tree can be naturally mapped to the tree describing molecular structure, as we demonstrate by providing a CFG that achieves that.

A context-free grammar thus allows us to efficiently represent molecules as non-binary trees with some constraints on tree structure. That efficiency matters for model training: RNN-style models have problems passing information over very long distances. The Transformer architecture we use doesn't have that problem, but its RAM requirements grow approximately as a square of the number of generation steps (in our case, rules needed to generate a given molecule). Finally, for any architecture, length of the sequence used to represent a molecule directly affects training time. 

Thus the average number of tokens needed to represent a molecule constrains the maximum size of molecules it's practical to produce.

Using a CFG allows us to combine a granular representation (yielding a combinatorial variety of structures), with explicit representation of particularly frequent or complex combinations (eg double aromatic cycles).

To generate guaranteed valid molecules, we introduce a new grammar (full listing in Appendix~\ref{app:grammar}) that respects atomic valences and explicitly introduces cycles.
\subsection{Respecting valences}
The initial rules are 
\begin{verbatim}
smiles -> nonH_bond
smiles -> initial_valence_1 bond
smiles -> initial_valence_2 double_bond
smiles -> initial_valence_3 triple_bond
\end{verbatim}
Any token ending in \verb|bond| represents an open bond to which some atom must be attached; \verb|nonH_bond| requires that that atom not be hydrogen (see Section~\ref{sec:hydrogen} on treatment of hydrogen). The molecule can then be grown via rules such as 
\begin{verbatim}
nonH_bond -> valence_2 bond
double_bond -> '=' valence_2
double_bond -> '=' valence_3 bond
double_bond -> '=' valence_4 double_bond
triple_bond -> '#' valence_3 
triple_bond -> '#' valence_4 bond
\end{verbatim}
So far it seems as if we could only produce linear molecules. However, the \verb|valence| tokens can expand to individual atoms as well as to structures that contain branches, for example
\begin{verbatim}
valence_4 -> 'C'
valence_3 -> 'N'
valence_3 -> valence_4 branch
valence_2 -> 'O'
valence_2 -> valence_3 branch
valence_2 -> valence_4 '(' double_bond ')'
\end{verbatim}
This allows us, so far, to generate arbitrary tree-shaped molecules, ie those without cycles.
\subsubsection{Implicit hydrogens}\label{sec:hydrogen}
A complication in the above approach is that the SMILES format allows for implicit hydrogens, ie any valence of any atom that's not specified is assumed to be occupied by hydrogen. Thus in effect, these hydrogen atoms are represented by empty strings, which is a problem for us as a CFG does not allow to expand a token into nothing (otherwise parsing would be much harder).

If the implicit hydrogen atoms were represented by for example lowercase \verb|h|, we could have grammar rules such as 
\begin{verbatim}
bond -> 'h'
bond -> nonH_bond
branch -> 'h'
branch -> '(' nonH_bond ')'
\end{verbatim}
If all we cared about was generating new SMILES strings, we could have used the above 4 rules and then inserted a post-processing step that replaces each \verb|h| with an empty string. However, we also want to be able to parse existing SMILES strings into our grammar, so we go with a two-step approach: we start with a grammar as described above, and then eliminate all occurrences of \verb|bond| by replacing them with either nothing or \verb|nonH_bond|, and similarly for \verb|branch|.

Thus for example, the rule
\begin{verbatim}
valence_3 -> valence_4 branch
\end{verbatim} is replaced by the two rules
\begin{verbatim}
valence_3 -> valence_4
valence_3 -> valence_4 '(' nonH_bond ')'
\end{verbatim}
\subsection{Cycles}
The different kinds of cycles are introduced by rules such as 
\begin{verbatim}
nonH_bond -> aliphatic_ring
aliphatic_ring -> valence_3_num cycle_bond
aliphatic_ring -> valence_4_num cycle_double_bond
valence_2 -> vertex_attached_ring
vertex_attached_ring -> valence_4_num '(' cycle_bond ')'
\end{verbatim}
The challenging bit here is the numbering - the two ends of a cycle must be marked with the same numeral, to know we should connect them; furthermore, SMILES reuses these numerals, so if the numeral \verb|1| was used in a cycle that's already closed, we can use it again for the next cycle we start. Thus cycle numeral management is a task that's quite hard to solve using just a context-free grammar.

We solve this by attaching additional properties to some nonterminal tokens. When we expand a nonterminal which has \verb|ring| as a substring, indicating a new cycle, we give it a unique cycle identifier, which is attached as a property to any tokens resulting in that expansion, which have substrings \verb|num| or \verb|cycle|. When these are in turn expanded, the cycle id is again propagated. The grammar rules for cycle propagation are formulated in a way that guarantees there are ever at most two tokens tagged by any one cycle ID, in fact exactly two until we expand them into numerals.
\subsubsection{Numeral assignment}
When a nonterminal is about to be expanded into a numeral, we know by construction that it has a cycle id. We have two cases to consider: firstly, there exists another token with this cycle id. In that case, we scan the token sequence so far to determine the lowest available numeral, and generate a mask that makes sure that specific numeral will be chosen; and store the pair (cycle id, numeral) in a cache. In the second case, when there are no tokens with this cycle id, we look up the numeral to use in the cache, by cycle id.
\subsubsection{Cycle size}
A cycle with only two atoms (same numeral attached to atoms that are already connected by a regular bond) is considered illegal in SMILES. To prevent these, we attach and propagate another property to nonterminals that are part of a cycle, namely ring size, incrementing it with each rule expansion. Cycle size masking then forbids the rules that would have led to premature cycle closure. As cycles of length more than 8 don't occur in the database (and our scoring function penalizes cycles with length more than 6), we also forbid cycle expansion beyond length 8.
\subsubsection{Cycle chaining}
Aliphatic cycles can form arbitrarily long, potentially branching chains, due to a pair of rules that allow starting a new cycle while constructing another:
\begin{verbatim}
cycle_bond -> aliphatic_ring_segment cycle_bond
aliphatic_ring_segment -> valence_3 '(' cycle_bond ')' valence_3_num
\end{verbatim}
In the first of these rules, the \verb|cycle_bond| token on the right hand side will get the same cycle id as that on the left hand side; while the \verb|aliphatic_ring_segment| token will trigger generation of a new cycle id, that will be propagated to the appropriate nonterminals on the right hand side of the second rule, in this case \verb|cycle_bond| and \verb|valence_3_num|.

This is just an example, a similar pattern is used to attach an aliphatic cycle to aromatic cycles described below. 
\subsubsection{Aromatic cycles}
Aromatic cycles and chains thereof must fulfil additional relationships between the number of atoms and the number of double bonds in each cycle. Solving this in a general way is beyond the scope of current work, we limit ourselves to rules that allow us to generate any single aromatic cycle of length 5 or 6, and common patterns of linked pairs of aromatic cycles, possibly with chains of aliphatic cycles attached. For example, here are some of the rules used in generating a coupled pair of aromatic rings of size 6 and 5:

\begin{verbatim}
double_aromatic_ring -> 'c' num1 aromatic_atom aromatic_atom 
                        aromatic_atom 'c' num 'n' num1 aromatic_atom aromatic_atom
                        aromatic_atom_num
aromatic_atom -> 'n'
aromatic_atom -> 'c'
aromatic_atom -> 'c' '(' nonH_bond ')'
aromatic_os -> 'o'
aromatic_os -> 's'
aromatic_atom_num -> 'c' num
\end{verbatim}
\subsection{Making sure the expansions terminate}
The final challenge we address is making sure the rule expansion terminates before the maximum allowed number of expansion rules. To do this, we define the concept of \emph{terminal distance} of a token, defined as the length of the shortest sequence of rules needed to transform that token into a sequence consisting only of terminals. 

We calculate that distance for all tokens reachable from the root token by means of the following algorithm: 
\begin{enumerate}
	\item Define a set $T$ of all tokens observed so far, and seed it with the root token \verb+smiles+. 
	\item Iterate over the elements $t$ of $T$. 
	\begin{enumerate}
		\item For each $t$ we firstly apply all applicable rules and add any new tokens generated thereby to $T$, seeding their terminal distance with 0 for terminals and $\infty$ for nonterminals.
		\item We then calculate the terminal distance of a token $t$ as one plus the minimum over all applicable rules of the sum of terminal distances of the tokens on the right hand side of the respective rule:
		\begin{equation}\label{eq:td}
		TD(t)=1 + \min_{r \in G: r.lhs = t}\left(\sum_{t' \in r.rhs} TD(t') \right)
		\end{equation}
	\end{enumerate}
\end{enumerate}

We repeat step 2. this until convergence, that is, until no new tokens are observed and the terminal distance for known tokens no longer changes after an iteration.

We then define the change in terminal distance made by a production rule as 
\begin{equation}\label{eq:dtd}
\Delta TD(r) =  \sum_{t' \in r.rhs} TD(t') - TD(r.lhs)
\end{equation}
By definition of the terminal distance, for every nonterminal $t$ there exists a production rule $r$ such that $r.lhs = t $ and $\Delta TD(r) = -1$.

Finally, we define the terminal distance of a token sequence as the sum of terminal distances of the individual tokens. 

We use the terminal distance concept to make sure the rule expansion terminates before the maximum number of steps, in the following manner: at each step, we consider the number $s$ of steps left and the terminal distance $td$ of the sequence generated so far. 

At each step, we make sure $td\le s$, using induction. First, we choose the maximum rule sequence length to be larger than the terminal distance of the root symbol. Second, at each rule selection step we consider all rules $r$ whose left hand side is the next nonterminal to expand, and only allow those where $\Delta TD(r) + td \le s-1$. That is a nonempty set because $\Delta TD(r')=-1$ for at least one applicable rule $r'$. Thus, by the time we run out of steps, that is, $s=0$, we know $td=0$, that is our token sequence consists only of nonterminals. 

Note that in our case, we have to apply this algorithm not to the tokens of our original grammar, but to the extended tokens (including cycle size information); and likewise use the production rules consisting of the original rules plus the propagation of cycle size information\footnote{This is an additional limit maximal cycle size, namely to make sure the number of extended tokens is finite}. This is because cycle masking forbids some rules (eg those that would create a cycle of two atoms) that would be allowed by our original grammar. Effectively, the rules for propagating additional information, along with the original CFG rules, induce a CFG on the space of extended tokens, and it is the terminal distance within that CFG that we must consider.
 
Also for computational efficiency reasons (to not multiply the number of tokens unnecessarily) we disregard the cycle id propety of the tokens when calculating terminal distance, as it only affects choice of numeral but not terminal distance. 

\subsection{Limits of context-free grammars}
Of the above modifications to a pure context-free grammar, cycle size masking could have been implemented by including the size value into the token string, and making a copy of each rule whose left hand side is that token, for every size value that occurs. This would increase the number of rules by a couple dozen, but stay within the limits of a CFG.

On the other hand, numeral choice and terminal masking do not appear possible to achieve using a CFG alone. However, these merely restrict the list of possible production rules at any step - but the resulting SMILES string is still valid according to the original CFG without these extra attributes. This is imporant because it allows us to use our original CFG to parse known molecules, eg from the ZINC database, to train our model on.

\subsection{Grammar conciseness and expressiveness}\label{sec:expressiveness}
Our approach leads to a more concise way of representing molecules using production rules, with an average 2.85 production rules per atom and 62.8 rules per molecule in the ZINC dataset, compared to average 5.46 rules per atom and 120.8 rules per molecule in the grammar used by \cite{kusner17}. This in turn allowed us to generate larger molecules (about 100 atoms for a typical molecule that uses the maximum number of steps, and over 400 atoms in special cases) on commodity hardware.

Our grammar can represent arbitrarily long, branching chains of aliphatic cycles, as well as single aromatic cycles with five or six atoms, pairs of aromatic cycles, and arbitrarily long branching chains of aliphatic cycles attached to aromatic ones.

Because our grammar is more restrictive than a fully generic SMILES grammar (to make it easier for us to generate guaranteed valid molecules), it can't represent all the molecules in the ZINC database, but rather a little over a third of the molecules, 92K out of 250K. Our grammar is also  expressive enough to represent the recent state-of-the art constructed molecules, such as those in \cite{kusner17} and \cite{jin18}. 

This approach can be extended to cover the whole ZINC database by introducing additional production rules - certainly by brute force, by adding to the grammar every unknown pattern that the parser encounters; and most likely also in a more elegant way. Extending our grammar to represent most or all molecules from that dataset is subject of ongoing work.

There is something of a tradeoff between expressiveness and guaranteed correctness - as one adds more features to the grammar, it becomes harder to maintain the guaranteed-correctness property. Fortunately, that property is not actually necessary for our approach (best-of-batch policy gradient) to work - it's sufficient for a large fraction, say 75\%, of each batch, to be valid for the optimization to succeed.

Future attempts to expand the grammar's expressiveness might make use of that by relaxing the guaranteed-correctness property if needed.

\section{Model choice}
\subsection{Reinforcement learning}
Most of the literature so far (\cite{Gomez-Bombarelli16},\cite{kusner17},\cite{jin18}) on molecule generation has focussed on training an autoencoder for the chosen molecule representation, using a database of known molecules such as ZINC, then doing Bayesian optimization over the latent space. 

We choose a different approach, treating molecule creation as a reinforcement learning problem. In our case, at each step the action taken by the network is choosing the next production rule; the state is the full sequence of the production rules so far; and the reward is nonzero for the step following molecule completion (ie when the sequence defined by applying the chosen production rules contains no more nonterminals), and zero for all other steps. Details of the reward specification are in Sections~\ref{sec:training}~and~\ref{sec:reward}.

The first step of the autoencoder/Bayesian optimization approach, training the autoencoder on a database of known molecules, typically served two objectives: firstly, ensuring that the model learns to produce chemically valid molecules (for the majority of models where that is not guaranteed by construction); secondly, ensuring that the molecules the model produced are `similar' to the molecules the model has been trained on.

While this approach has many desirable traits, it also has disadvantages: firstly, once the autoencoder is trained, its notion of `similarity to known good molecules' is cast in stone, and no subsequent optimization over the latent space would allow it to produce, for example, models with similar structure but substantially larger than the ones it was trained on. In contrast, casting molecule generation as a reinforcement learning problem allows us to treat `similarity to known good molecules' as the gradual concept that it is, for example (as we do) by including a variably-weighted loss term penalizing the coefficient distance from a version of the model trained off-policy on known-good molecules. This will allow us to search over a greater space of possible molecules, while maintaining control over the degree of similarity to the known-good ones.

Secondly, although the molecules in the latent space neighborhood of a certain molecule are guaranteed to be `similar' to it in some sense, that neighborhood is not guaranteed to contain \emph{all} the `similar' molecules we might be interested in. In comparison, in a reinforcement learning problem we can explicitly specify any proximity metric we desire, and simply include it in the reward. 

Thirdly, learning a latent space representation of the space of all the molecules in the database is a reasonably hard task, which is not necessary if all we want is to generate molecules that maximize some metric and are similar to existing ones. We conjecture that simply teaching a model to have a high probability of producing those molecules is a computationally cheaper task, which can be used to achieve the same aim via anchoring (see below).

Finally, given the extremely non-Euclidean structure of the space of all valid molecules, it seems promising to conduct the search for optimal molecules directly on a representation that seems close to it, in our case the sequence of grammar production rules tailored to molecular structure, rather than on the Euclidean latent space. 
\subsection{Architecture}
As the molecular properties we seek to optimize are essentially nonlocal, and because a grammar-rule representation of a branching tree means that even atoms that are nearby on the tree can end up generated by rules that are far apart in the rules sequence, we use the Transformer architecture \cite{Transformer17}, chosen because any two items in the sequence are directly linked by its attention mechanism. Our implementation of the Transformer decoder outputs one vector of logits per call (for each molecule in the batch), to be used in choosing the next production rule for each sequence in the batch; and omits the calculation steps that use the sequence generated by the encoder as it doesn't exist in our caase.

We use 6 layers, 6 heads, $d_k=d_v=16$, $d_{model}=128$, $d_{inner\_hid}=256$.

Prior to calculating the log-softmax to be used for sampling the next grammar rule, we calculate a mask as described in Section~\ref{sec:valid_smiles} and subtract 1e6 from the logits at the indices forbidden by the mask. 
\subsection{Training}\label{sec:training}
We use same maximal sequence length as \cite{kusner17}, 277.
That allows us a maximum batch size of 40 on an AWS p2.xlarge instance. 

In the first stage, we train the model off-policy on 92K molecules from the ZINC database  \cite{Weininger88} that can be represented by our grammar.  In order to reward the model for producing molecules similar to the ones it observes, we use simple policy gradient loss (\ref{eq:pgloss}). Here $s$ goes over the model steps, and $\pi_s\left(r_s\right)$ is the model-produced probability (after applying all masking) of choosing rule number $r_s$ at step $s$.
\begin{equation}\label{eq:pgloss}
loss = -\sum_{s=0}^S \log\left(\pi_s\left(r_s\right)\right)
\end{equation}
We optimize in batches of 40 molecules, using the average loss for the batch, and Adam optimizer with a learning rate of 1e-4. After each batch, we do an on-policy simulation of a batch of 40 molecules with the same loss function, to judge convergence - but without updating the model coefficients.

To judge convergence, we compare the distribution of log probability, logP, SA score, and the number of aromatic cycles in the molecules produced by the model to those in the database, and find that they have converged after 15 epochs of training (30K batches, about one week hours on a p2.xlarge). We take the model thus trained as our base model.


We then proceed to on-policy training using best-of-batch policy gradient with base model anchoring. That is, our loss is the log-likelihood \emph{for the best-scoring molecule in the batch} plus the L2-distance between the coefficients of the current model and the base model, multiplied by a weight $w_a$(\ref{eq:pgloss_anchor}). The purpose of the second term (`anchoring') is to prevent catastrophic forgetting, and to control the degree of similarity of optimized model to the base model (and thus to the molecules in the database).

\begin{equation}\label{eq:pgloss_anchor}
loss =- \sum_{s=0}^S \log\left(\pi_s\left(r_s\right)\right) + w_a\left| \vec p - \vec p_{base} \right|_2^2
\end{equation}

The reward function is used to choose the best molecule of the batch, but its value for the best molecule does not affect the loss function, which remains a simple log likelihood for the best molecule.

We found that using the best molecule in the batch, rather than the batch average, during the second phase was crucial for successful optimization - without it, as soon as the search found a somewhat successful molecule, the optimization tended to converge to always producing that molecule; while taking best-of-batch enabled continued exploration.

\subsection{Optimization and the reward function}\label{sec:reward}
We optimize the metric used in prior literature (\cite{Gomez-Bombarelli16}, \cite{kusner17}, \cite{jin18}), namely logP minus the synthethic accessibility score, penalized for cycles of more than 6 atoms. Each of those three terms is evaluated for each of the molecules in the ZINC dataset using code shared by \cite{kusner17}, and their mean and standard deviation are calculated. These are used to normalize each of the 3 score constituents, before adding them up to calculate a molecule's score.

The overall reward function is thus represented by (\ref{eq:reward}). 
\begin{equation}\label{eq:reward}
R = logP_{norm} + SA_{norm} + C_{norm}
\end{equation}

For reasons explained in Section~\ref{sec:results}, we use an extended version of (\ref{eq:reward}), namely (\ref{eq:reward2}).
\begin{equation}\label{eq:reward2}
R = logP_{norm} + SA_{norm} + C_{norm} + w_{SA} \min\left( SA_{norm}, 0\right) - w_{ac} \min\left( num\_aromatic\_cycles -5, 0\right)
\end{equation}
Here the first three terms on the right are the standard scoring function used in the literature, and the final two terms allow us to penalize low SA scores and too large a number of aromatic cycles in a molecule.


\section{Results}\label{sec:results}
Depending on how harshly we penalize the distance to the base model (choice of $w_a$), our model produces a range of molecules with scores substantially exceeding state-of-the-art, with varying sizes, as shown in Table~\ref{tab:table1} and discussed in detail below.

\begin{figure}[ht]
	\centering
	\includegraphics[width=0.8\textwidth]{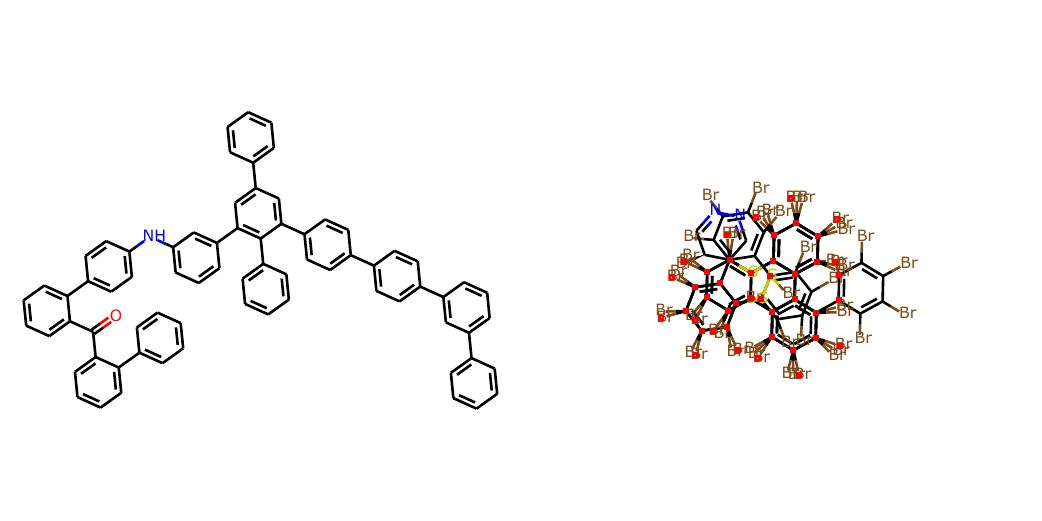}
	\caption{Top molecule with weak anchoring and SA penalty (left) and no anchoring (right)}	\label{fig:optim}
\end{figure}

\begin{table}[ht]
	\begin{center}
		
		\begin{tabular}{|l|r|r|r|r|} 
			\hline
			Source & Score & SA & Norm.SA & Aromatic rings\\
			\hline
			Average over ZINC database&0.0&-2.82& 0.0 &1.85\\
			Standard deviation over ZINC database&2.07&0.77&1.0&0.97\\\hline
			Kusner et al. & 2.93 & -1.93 & 1.35 & 1 \\
			Jin et al. & 5.30 & -2.33 & 0.87 & 5 \\\hline
			{\bf Strong Anchor, SA penalty 1st} &{\bf 5.68} & -1.96 & 1.31 & 5 \\
			Strong Anchor, SA penalty 2nd & 5.60 & -2.62 & 0.52 & 4 \\
			Strong Anchor, SA penalty 3rd & 5.33 & -2.87 & 0.21 & 5 \\\hline
			Weak Anchor, SA and aromatic cycle penalty 1st & 9.48 & -2.83 & 0.27 & 7 \\
			Weak Anchor, SA and aromatic cycle penalty 2nd & 8.73 & -2.98 & 0.08 & 5 \\
			Weak Anchor, SA and aromatic cycle penalty 3rd & 8.44 & -2.67 & 0.46 & 6 \\\hline
			Weak Anchor, SA penalty & 12.25 & -2.98 & 0.08 & 12 \\
			{\bf Unconstrained}& {\bf 46.45} & -1.00 & 2.46 & 13 \\
			\hline
		\end{tabular}
		\caption{Statistics of optimized molecules (higher scores are better)}\label{tab:table1}
	\end{center}
\end{table}
When trying to optimize  (\ref{eq:reward}) without constraning similarity to known molecules ($w_{a} =0$) and using the original scoring function from the literature (that is, $w_{SA}=0$ and  $w_{ac}=0$), we've found that the optimization tried to get as many aromatic cycles as possible, leading to a large logP but very low SA score, and then attach a large number of halogen atoms to these to bring the SA score back up (see Figure~\ref{fig:optim}, right). While this allowed for an order of magnitude increase in the molecule score (`Unconstrained' in Table~\ref{tab:table1}), such molecules don't appear to be realistic candidates for having useful properties. 

We next added weak anchoring ($w_{a} =1e8$) and a penalty on negative normalized SA score ($w_{SA}=20$). The result is shown in Table~\ref{tab:table1} as `Weak Anchor, SA penalty' and in Fig~\ref{fig:optim}, left. We see that the introduced constraints temper but don't entirely remove the tendency of the optimization to blindly maximize the score by adding a lot of aromatic cycles.

\begin{figure}[ht]
	\centering
	\includegraphics[width=\textwidth]{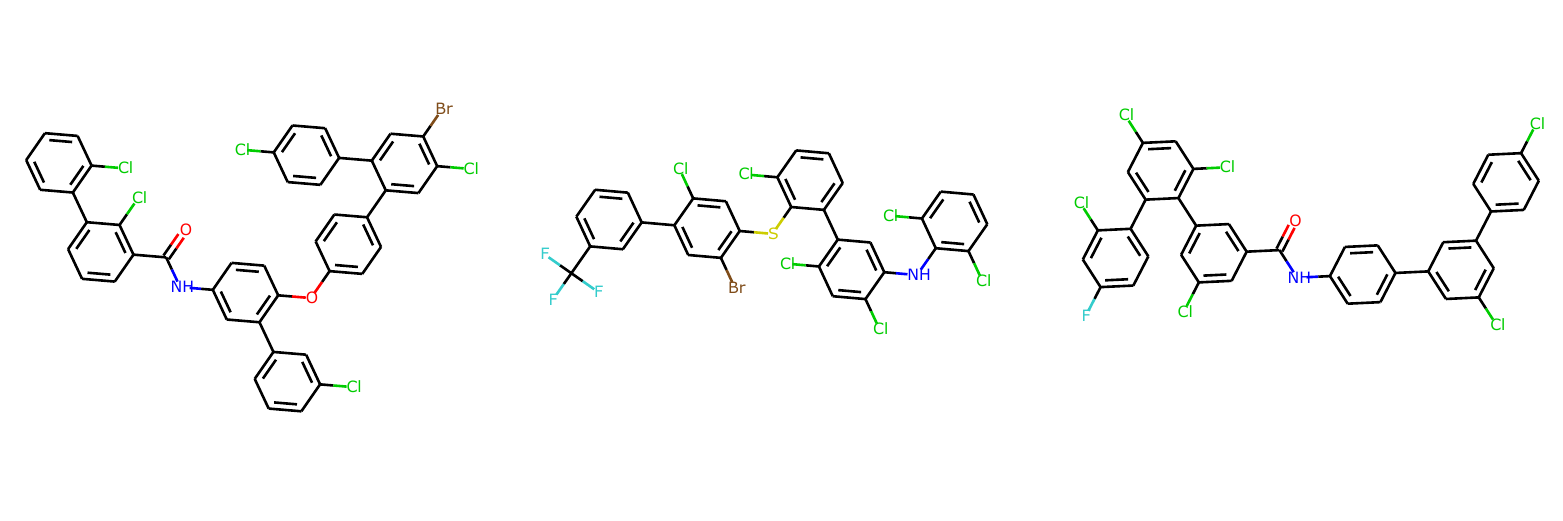}
	\caption{Top 3 molecules with weak anchoring, and SA and aromatic cycle penalty}\label{fig:weak}
\end{figure}

To counteract that, we add a penalty on the number of aromatic cycles exceeding 5 ($w_{ac}=5$). The three top-scoring molecules are shown in Table~\ref{tab:table1} as `Weak Anchor, SA penalty and aromatic cycle penalty', and in Figure \ref{fig:weak}. We see that even after constraining the number of aromatic cycles, we can achieve scores well in excess of 8, and a larger variety of atoms and structures than in the previous cases. However, we still see many more halogens than in a typical molecule from the ZINC database.

\begin{figure}[ht]
	\centering
	\includegraphics[width=\textwidth]{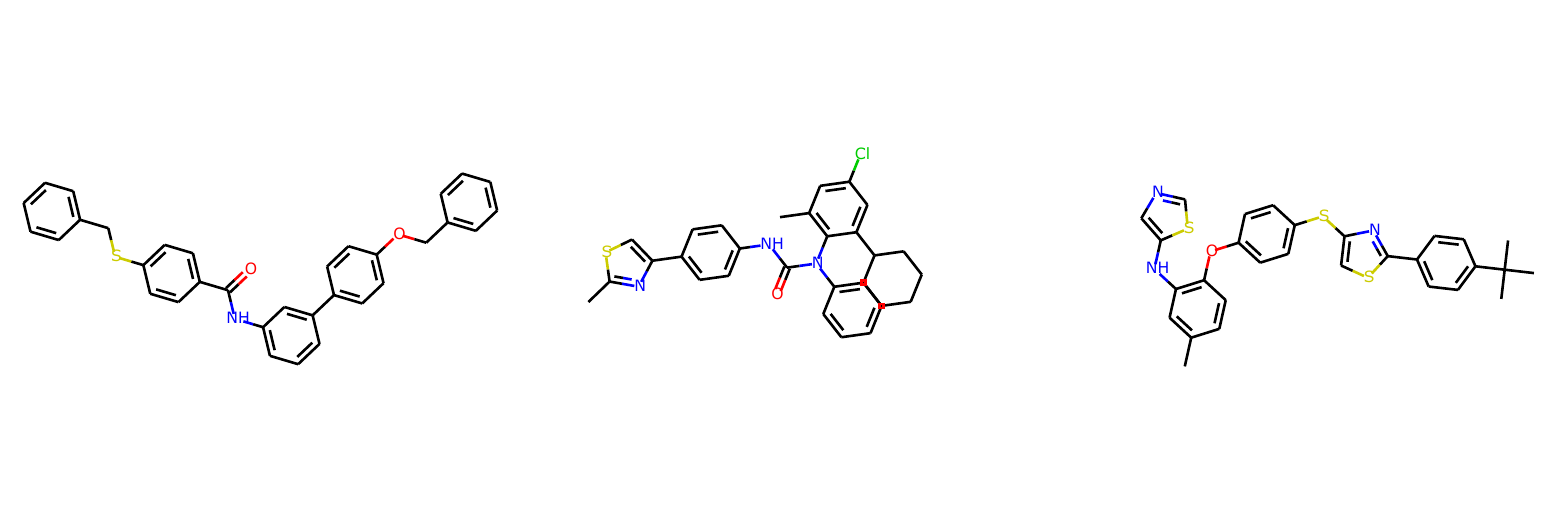}
	\caption{Top 3 molecules with strong anchoring and SA penalty}\label{fig:strong}
\end{figure}

For a final set of simulations, we increase the anchoring weight to $w_{a} =2e8$, keep the SA penalty as before, and remove the aromatic cycle penalty. The three top-scoring molecules are shown in Table~\ref{tab:table1} as `Strong Anchor, SA penalty', and in Figure~\ref{fig:strong}. We see that these molecules appear much more similar to the ZINC database - in addition to the aromatic cycles of 6 carbon atoms we see an aliphatic cycle, aromatic 5-cycles containing sulphur and nitrogen, and just one chlorine atom. All of these molecules' scores exceed those of the top molecule of \cite{jin18}.


\section{Conclusion}
A major contribution of \cite{jin18} was regarding the molecule as a junction tree, with any cycles represented as nodes on par with non-cycle atoms. This made it possible to create a model guaranteed to generate valid molecules. However, their implementation was comparatively complex and also required all possible cycles to be enumerated in a dictionary upfront. 

Here we have shown that the same idea can be implemented in a simpler fashion, using a custom context-free grammar, combined with some additional masking for choosing the next production rule. 

A properly constructed context-free grammar gives us the best of both worlds: the ability, with a small number of production rules, to represent all possible valid combinations of atoms to build a given structure, as in \cite{kusner17}; and at the same time, the ability to explicitly specify complex structures (eg linked aromatic rings) upfront using a single rule per structure, similarly to \cite{jin18}, while letting further rule expansions supply the detail. All that is done as part of a single grammar, with the decomposition of a molecule into a production rule sequence being done using a standard CFG parser, and the generation of new molecules being done by any model able to recursively produce a sequence of tokens (we chose the Transformer because of its nonlocal information propagation, but an RNN stack could have been used as well).

To solve some of the shortcomings of a purely CFG-based approach, we propagated additional information with the tokens during rule expansion, and used it to limit the set of allowed production rules at each step. 

This approach, combining a custom CFG with additional rule masking that uses non-local information, is applicable beyond molecules, to any domain where we need to optimize complex graph structures under a mixture of local and nonlocal constraints: the local constraints can be taken care of in the CFG design, and the nonlocal ones enforced via additional masking.

Finally, we show that molecule optimization can be naturally cast as a reinforcement learning problem, and the state-of-the-art results we get even with the very basic policy gradient method suggest further scope for improvement using more advanced approaches, for example methods based on tree search.

\bibliography{bibliography}
\bibliographystyle{apalike}

\appendix
\section{Full Listing of the Grammar}\label{app:grammar}
\tiny
\begin{verbatim}
smiles -> nonH_bond
smiles -> initial_valence_1
smiles -> initial_valence_1 nonH_bond
smiles -> initial_valence_2 double_bond
smiles -> initial_valence_3 triple_bond
initial_valence_1 -> 'F'
initial_valence_1 -> 'Cl'
initial_valence_1 -> 'Br'
initial_valence_1 -> 'I'
initial_valence_1 -> '[' 'O' '-' ']'
initial_valence_1 -> '[' 'N' 'H' '3' '+' ']'
initial_valence_2 -> 'O'
initial_valence_2 -> 'S'
initial_valence_3 -> '[' 'C' '@' 'H' ']'
initial_valence_3 -> '[' 'C' '@' '@' 'H' ']'
initial_valence_3 -> 'N'
initial_valence_3 -> '[' 'N' 'H' '+' ']'
nonH_bond -> valence_1
nonH_bond -> valence_2
nonH_bond -> valence_2 nonH_bond
nonH_bond -> valence_3 double_bond
nonH_bond -> valence_4 triple_bond
double_bond -> '=' valence_2
double_bond -> '=' valence_3
double_bond -> '=' valence_3 nonH_bond
double_bond -> '=' valence_4 double_bond
triple_bond -> '#' valence_3
triple_bond -> '#' valence_4
triple_bond -> '#' valence_4 nonH_bond
valence_4 -> 'C'
valence_4 -> '[' 'C' '@' ']'
valence_4 -> '[' 'C' '@' '@' ']'
valence_4 -> '[' 'N' '+' ']'
valence_3 -> '[' 'C' '@' 'H' ']'
valence_3 -> '[' 'C' '@' '@' 'H' ']'
valence_3 -> 'N'
valence_3 -> '[' 'N' 'H' '+' ']'
valence_3 -> valence_4
valence_3 -> valence_4 '(' nonH_bond ')'
valence_2 -> 'O'
valence_2 -> 'S'
valence_2 -> 'S' '(' '=' 'O' ')' '(' '=' 'O' ')'
valence_2 -> valence_3
valence_2 -> valence_3 '(' nonH_bond ')'
valence_2 -> valence_4 '(' double_bond ')'
valence_1 -> 'F'
valence_1 -> 'Cl'
valence_1 -> 'Br'
valence_1 -> 'I'
valence_1 -> '[' 'O' '-' ']'
valence_1 -> '[' 'N' 'H' '3' '+' ']'
valence_1 -> valence_2
valence_1 -> valence_2 '(' nonH_bond ')'
valence_1 -> valence_3 '(' double_bond ')'
valence_1 -> valence_4 '(' triple_bond ')'
nonH_bond -> valence_2 slash valence_3 '=' valence_3 slash valence_2
slash -> '/'
slash -> '\'
nonH_bond -> aliphatic_ring
aliphatic_ring -> valence_3_num cycle_bond
aliphatic_ring -> valence_4_num cycle_double_bond
valence_2 -> vertex_attached_ring
vertex_attached_ring -> valence_4_num '(' cycle_bond ')'
cycle_bond -> valence_2 cycle_bond
cycle_bond -> valence_3 cycle_double_bond
cycle_double_bond -> '=' valence_3 cycle_bond
cycle_bond -> valence_2_num
cycle_double_bond -> '=' valence_3_num
nonH_bond -> aliphatic_ring_segment
nonH_bond -> aliphatic_ring_segment nonH_bond
cycle_bond -> aliphatic_ring_segment cycle_bond
aliphatic_ring_segment -> valence_3 '(' cycle_bond ')' valence_3_num
aliphatic_ring_segment -> valence_4 '(' cycle_bond ')' '=' valence_4_num
aliphatic_ring_segment -> valence_4 '(' cycle_double_bond ')' valence_3_num
nonH_bond -> aromatic_ring_5
nonH_bond -> aromatic_ring_6
starting_aromatic_c_num -> 'c' num
aromatic_atom -> 'n'
aromatic_atom -> 'c'
aromatic_atom -> 'c' '(' nonH_bond ')'
aromatic_os -> 'o'
aromatic_os -> 's'
aromatic_os -> 'n' '(' nonH_bond ')'
aromatic_os -> '[' 'n' 'H' ']'
aromatic_atom_num -> 'n' num
aromatic_atom_num -> 'c' num
aromatic_atom_num -> 'c' num nonH_bond
aromatic_os_num -> 'o' num
aromatic_os_num -> 's' num
aromatic_os_num -> 'n' num nonH_bond
nonH_bond -> double_aromatic_ring
double_aromatic_ring -> 'c' num1 aromatic_atom aromatic_atom aromatic_atom 'c' num 'c' num1 aromatic_atom aromatic_atom aromatic_atom aromatic_atom_num
double_aromatic_ring -> 'c' num1 aromatic_atom aromatic_atom aromatic_atom 'c' num 'n' num1 aromatic_atom aromatic_atom aromatic_atom_num
double_aromatic_ring -> 'c' num1 aromatic_atom aromatic_atom aromatic_atom 'n' num 'c' num1 aromatic_atom aromatic_atom aromatic_atom_num
aromatic_ring_6 -> starting_aromatic_c_num aromatic_atom full_aromatic_segment aromatic_atom aromatic_atom_num
aromatic_ring_6 -> starting_aromatic_c_num full_aromatic_segment full_aromatic_segment aromatic_atom_num
aromatic_ring_5 -> starting_aromatic_c_num aromatic_os full_aromatic_segment aromatic_atom_num
aromatic_ring_5 -> starting_aromatic_c_num aromatic_atom aromatic_os aromatic_atom aromatic_atom_num
aromatic_ring_5 -> starting_aromatic_c_num full_aromatic_segment aromatic_os aromatic_atom_num
aromatic_ring_5 -> starting_aromatic_c_num full_aromatic_segment aromatic_atom aromatic_os_num
aromatic_ring_5 -> starting_aromatic_c_num aromatic_atom full_aromatic_segment aromatic_os_num
full_aromatic_segment -> aromatic_atom aromatic_atom
aromatic_os -> side_aliphatic_ring
side_aliphatic_ring -> 'c' num '(' cycle_bond ')'
full_aromatic_segment -> side_aliphatic_ring_segment
side_aliphatic_ring_segment -> 'c' num 'c' '(' cycle_bond ')'
side_aliphatic_ring_segment -> 'c' '(' cycle_bond ')' 'c' num
valence_4_num -> 'C' num
valence_4_num -> '[' 'C' '@' ']' num
valence_4_num -> '[' 'C' '@' '@' ']' num
valence_4_num -> '[' 'N' '+' ']' num
valence_3_num -> '[' 'C' '@' 'H' ']' num
valence_3_num -> '[' 'C' '@' '@' 'H' ']' num
valence_3_num -> 'N' num
valence_3_num -> '[' 'N' 'H' '+' ']' num
valence_3_num -> valence_4_num
valence_3_num -> valence_4_num '(' nonH_bond ')'
valence_2_num -> 'O' num
valence_2_num -> 'S' num
valence_2_num -> 'S' num '(' '=' 'O' ')' '(' '=' 'O' ')'
valence_2_num -> valence_3_num
valence_2_num -> valence_3_num '(' nonH_bond ')'
valence_2_num -> valence_4_num '(' double_bond ')'
num -> '1'
num1 -> '1'
num -> '2'
num1 -> '2'
num -> '3'
num1 -> '3'
num -> '4'
num1 -> '4'
num -> '5'
num1 -> '5'
num -> '6'
num1 -> '6'
num -> '7'
num1 -> '7'
num -> '8'
num1 -> '8'
num -> '9'
num1 -> '9'
num -> '%10'
num1 -> '%10'
num -> '%11'
num1 -> '%11'
...
num -> '%49'
num1 -> '%49'
\end{verbatim}

\end{document}